\documentclass{ws-procs11x85}
\usepackage{ws-procs-thm}           

\usepackage{booktabs}
\usepackage{graphicx}
\usepackage{array}
\usepackage{enumitem}
\usepackage{amsmath}
\usepackage{multirow}
\usepackage{hyperref}
\usepackage{xcolor}
\usepackage{longtable}
\usepackage[load-configurations=version-1]{siunitx}
\usepackage{comment}  

\definecolor{website_color}{HTML}{E91E63}

\begin{document}


\def\BENCHMARK{ReasonKnee-Bench}
\def\DATASET{3DReasonKnee}

\title{3DReasonKnee: Advancing Grounded Reasoning in \\ Medical Vision Language Models}

\author{%
Sraavya Sambara$^{1*}$, Sung Eun Kim$^{1,2*}$, Xiaoman Zhang$^{1}$, Luyang Luo$^{1}$ \\ 
Shreya Johri$^{1}$, Mohammed Baharoon$^{1}$, Du Hyun Ro$^{2}$, Pranav Rajpurkar$^{1}$
}

\address{
$^1$Department of Biomedical Informatics, Harvard Medical School, Boston, MA, USA \\
$^2$Seoul National University Hospital, Seoul, South Korea}

\copyrightinfo{*These authors contributed equally to this work. \\ \copyright\ 2025 The Authors. Open Access chapter published by World Scientific Publishing Company and distributed under the terms of the Creative Commons Attribution Non-Commercial (CC BY-NC) 4.0 License.}

\begin{abstract}


Current Vision-Language Models (VLMs) struggle to ground anatomical regions in 3D medical images and reason about them in a step-by-step manner---a key requirement of real-world diagnostic assessment. This ability is essential for aligning model outputs with the diagnostic workflows clinicians use in practice, enabling trustworthy clinician-AI collaboration. Existing 3D datasets provide localization labels, but none support this ``grounded reasoning'' ability. To address this gap, we introduce \textbf{\DATASET}, the first 3D grounded reasoning dataset for medical images, which provides 494k high-quality quintuples derived from 7,970 3D knee MRI volumes. Each quintuple includes: (1) the 3D MRI volume, (2) a diagnostic question targeting a specific anatomical region (3) a 3D bounding box localizing the relevant anatomical structures, (4) clinician-generated diagnostic reasoning steps that explicitly detail the 3D reasoning process, and (5) structured severity assessments for the relevant anatomical region. The meticulous creation and validation of \DATASET, involving over 450 hours of expert clinician time for manually segmenting MRIs and generating reasoning chains, ensures its superior quality and clinical relevance. 
We establish \textbf{\BENCHMARK} to evaluate localization and diagnostic accuracy, providing novel insight into VLM ability to perform grounding and severity assessment across diverse anatomical regions and diagnostic inquiries. We benchmark five state-of-the-art VLMs, providing baseline performance for \BENCHMARK.
By providing this unique resource of expert-annotated 3D reasoning pathways, \DATASET~ serves as a repository of orthopedic surgeons’ diagnostic expertise and offers a vital testbed for advancing multimodal medical AI systems towards 3D, clinically aligned, localized decision-making capabilities. The dataset can be found in HuggingFace: \href{https://huggingface.co/datasets/rajpurkarlab/3DReasonKnee}{\textcolor{blue}{rajpurkarlab/3DReasonKnee}}.
\end{abstract}

\section{Introduction}

\begin{figure}[!th]
    \centering
    \includegraphics[width=1\linewidth]{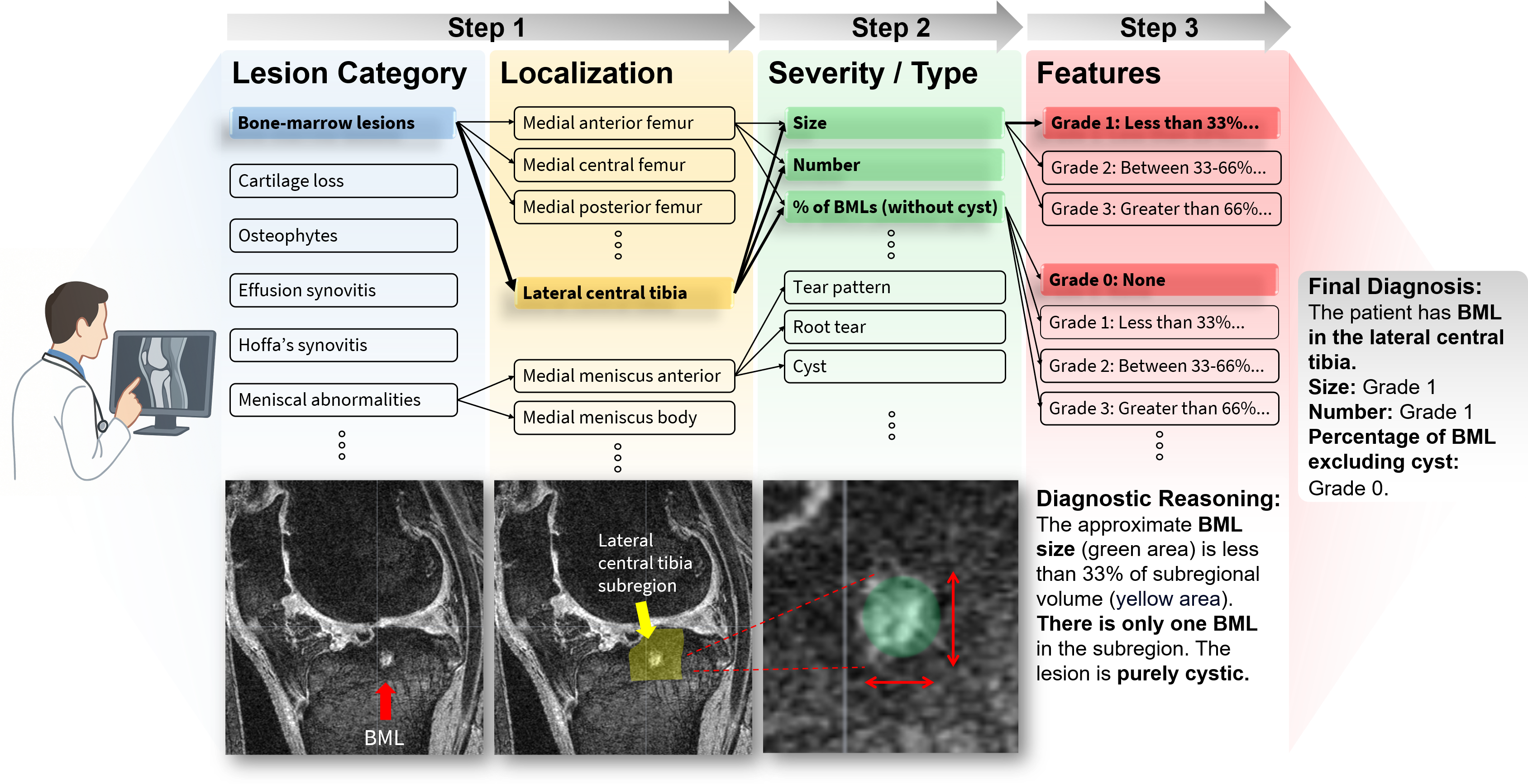}
    \caption{Region-First Reasoning Workflow in Clinician Knee MRI Assessment. This schema illustrates the step-wise diagnostic process employed by clinicians, involving (Step 1) identification of lesion category and localization to specific anatomical subregions, (Step 2) assessment of severity and lesion type within the localized region, and (Step 3) characterization of specific features.} 
    \label{fig:schema}
\end{figure}

\textbf{Medical VLMs Lack Fine-Grained 3D Grounded Reasoning.} Recent progress in Medical Vision-Language models (VLMs) has demonstrated remarkable capabilities in versatile tasks requiring integrated reasoning across modalities, with models achieving state-of-the-art performance on visual question answering, report generation, and other global understanding tasks~\cite{li2025visual,zhang2023multimodal}. However, these models still fall short of clinician-level performance when confronted with problems demanding fine-grained diagnostic reasoning and assessment. Clinicians typically employ a region-first workflow (Figure~\ref{fig:schema}), pinpointing relevant anatomical subregions, assessing them for abnormalities such as lesions or structural changes, and finally assigning structured severity grades based on established clinical criteria. For instance, the MRI Osteoarthritis Knee Score (MOAKS) framework, as well as other semi-quantitative grading systems, necessitates this level of fine-grained analysis to evaluate specific knee structures and assign severity grades to features like bone marrow lesions and cartilage damage. \cite{Roemer2011HOAMS, Haugen2015HOAMRIS} This "grounded" understanding of pathology within localized regions is crucial for accurate diagnosis, prognosis, and treatment planning in clinical settings. Bridging this gap between clinician reasoning and model intelligence is critical for enabling trustworthy clinical decision support and seamless integration of AI into diagnostic workflows.

\noindent \textbf{3D Grounded Reasoning is Not Well-Supported by Existing Datasets.} Advances in Chain-of-thought (CoT) prompting have proven effective in enabling models to break down complex problems into interpretable, step-by-step solution processes~\cite{wei2022chain,zhang2024chain}. In 3D medical image analysis, diagnostic reasoning represents a structured cognitive process that extends far beyond simple pattern recognition. It requires precise spatial understanding across multiple planes to properly evaluate volumetric structures. However, instilling such reasoning capabilities in Medical VLMs for 3D medical data is hampered by a lack of suitable training data. While a growing number of datasets are facilitating the development of medical VLMs, such as MultiMedEval~\cite {royer2024multimedeval} for diverse evaluation and MedTrinity-25M~\cite{xie2025medtrinitym} for large-scale pretraining, these resources often focus on 2D images with reasoning traces or 3D corpora emphasizing localization without detailed diagnostic rationale  (Table~\ref{tab:vlm_datasets}). A significant gap remains in datasets that facilitate grounded reasoning on 3D medical images, particularly those combining volumetric data, integrated voxel-level localization, and expert-annotated reasoning chains.

\noindent  \textbf{Introducing \DATASET~ to Advance 3D Grounded Reasoning.} Our work introduces \DATASET, the first 3D grounded reasoning dataset for medical images, comprising 494k high-quality quintuples spanning 7,970 unique MRIs. Each quintuple includes: (1) a 3D knee MRI volume, (2) a diagnostic question targeting a specific anatomical region, (3) a 3D bounding box localizing the relevant anatomical structures, (4) clinician-generated diagnostic reasoning steps detailing the 3D reasoning process, and (5) structured severity assessments for the relevant anatomical region. We frame 62 distinct clinical instruction tuning questions and provide volumetric bounding boxes for 22 distinct anatomical regions, drawing from the MOAKS framework. The creation of \DATASET~ involved over 450 clinician hours dedicated to manually labeling MRI data and generating detailed reasoning traces. In doing so, \DATASET~ provides a clinically aligned resource that supports interpretable model development and enables rigorous evaluation of model performance on narrowly defined diagnostic tasks.
To leverage this rich data, we establish \BENCHMARK~to rigorously evaluate both localization and diagnostic accuracy in VLMs. Our evaluation metrics include diagnostic accuracy assessment and 3D IoU for the model-generated localization bounding boxes, providing new insights into VLM ability to perform grounding and severity assessment across diverse anatomical regions and diagnostic inquiries. We emphasize the significant potential of \DATASET, with its expert-annotated reasoning chains and validated localization labels serving as a valuable clinical gold standard, to drive advancements in VLM grounded reasoning capabilities. This work lays the foundation for VLMs that better reflect how clinicians localize and reason about findings in practice, an important step toward their reliable and interpretable use in diagnostic care.

\begin{table}[!t]
\centering
\tbl{\small Representative datasets and benchmarks for medical VLMs.  Dim.~denotes dimensionality (2D vs.\ 3D).  Grounding indicates the presence of spatial annotations (bounding boxes, masks, or voxel labels); refers to expert-authored diagnostic reasoning steps or structured interpretation pathways that mirror how clinicians synthesize visual findings into diagnostic conclusions}{
\begin{tabular}{lcccccc}
\toprule
Dataset & Dim. & Modality & Size & Grounding & Reasoning \\ 
\midrule
Path‐VQA & 2D & pathology images & 32 k samples & $\times$ & $\times$ \\
Chest ImaGenome & 2D & chest X-ray & 242 k samples & \checkmark{} & $\times$ \\
PadChest‐GR & 2D & chest X-ray & 4,555 studies & \checkmark{} & $\times$ \\
PubMedVision & 2D & biomedical figures & 1.3 M samples & $\times$ & $\times$ \\
RadRBench-CXR & 2D & chest X-ray &  59K samples & $\times$ & \checkmark{}   \\
GMAI-Reasoning10K & 2D & mixed modalities  & 10K samples & $\times$ & \checkmark{}   \\
MedTrinity-25M & 2D & mixed modalities & 25M samples &  \checkmark{} & \checkmark{} \\
\midrule
M3D‐RefSeg & 3D & CT & 210 vols & \checkmark{} & $\times$ \\
MedMD RP3D & 3D & CT/MR & 51 k vols & $\times$ & $\times$ \\
RadGenome-Chest CT & 3D & CT & 1.3 M samples & \checkmark{} & $\times$ \\
\texttt{3D ReasonKnee } & 3D & knee MRI & 494 k samples & \checkmark{} & \checkmark{} \\
\bottomrule
\end{tabular}
}
\label{tab:vlm_datasets}
\end{table}

\section{Related Work}

\paragraph{Existing Datasets and Benchmarks for Medical VLMs.}
A growing number of datasets and benchmarks are facilitating the development of medical vision-language models.
MultiMedEval~\cite{royer2024multimedeval} offers a consolidated evaluation framework across various tasks and domains, with its VQA component utilizing 2D datasets such as PMC-VQA~\cite{zhang2024development}, Path-VQA~\cite{naseem2022vision}, SLAKE~\cite{liu2021slake}, and VQA-Rad~\cite{lau2018dataset}. MedTrinity-25M~\cite{xie2025medtrinitym} introduces a large-scale multimodal dataset of 25 million image-text pairs for medical vision-language pretraining. GMAI-Reasoning10K~\cite{su2025gmai} provides 10,000 medical reasoning examples with detailed CoT explanations for each VQA pair. RadRBench-CXR~\cite{fan2025chestx} offers 59K visual question answering samples with 301K clinically validated reasoning steps. OpenBiomedVid~\cite{thapa2025well} introduces a biomedical
video instruction tuning dataset from public educational videos.
MedMD~\cite{wu2023towards} includes 3D images and corresponding
captions in its RP3D subset. Similarly, 
M3D-Data \cite{bai2024m3d} introduces image-text pairs and instruction-response pairs tailored for 3D medical tasks, including voxel-level grounding in its M3D-RefSeg subset.
RadGenome-ChestCT~\cite{zhang2024radgenome} presents 665K multi-granularity grounded reports and 1.3M VQA pairs of 25,692 3D chest CT volumes from CT-RATE~\cite{hamamci2024foundation}.
Despite these advances, there remains a significant gap in datasets that facilitate grounded reasoning on 3D medical images, particularly those that combine volumetric data with expert-annotated reasoning chains. Our proposed 3D ReasonKnee addresses this gap by providing the largest 3D MRI dataset with integrated voxel-level localization, expert CoT rationales, and structured severity grades.


\paragraph{Medical VLMs: Capabilities and Gaps.}  The application of VLMs in the medical domain has led to models like MedVersa~\cite{zhou2024generalist}, which explores versatile instruction following tasks for both 2D and 3D medical images. LLaVA-Med~\cite{li2023llava} adapts the LLaVA framework for biomedical images, but predominantly on 2D data and without explicit reasoning steps. MAIRA-2~\cite{bannur2024maira} focuses on grounded report generation, mainly for chest X-rays, underscoring the need for more spatially annotated datasets supporting grounding in medical imaging. While models like Med3DVLM~\cite{xin2025med3dvlm} and M3D-LaMed~\cite{bai2024m3d} have been developed for 3D medical image analysis, leveraging datasets like M3D, they remain limited in diagnosis with vision-language reasoning. Other notable medical VLMs such as RadFM~\cite{wu2023towards}, PMC-VQA~\cite{zhang2024development}, MedPalm M~\cite{tu2024towards}, and BiomedGPT~
\cite{zhang2024generalist} have advanced medical image understanding and report generation, but generally lack robust support for fine-grained 3D grounded reasoning with associated expert rationales.
Therefore, the existing landscape of medical VLM datasets and models reveals a significant gap in resources that facilitate grounded reasoning on 3D medical images. \DATASET~aims to bridge this gap by introducing the first large-scale 3D MRI dataset for knee imaging that simultaneously offers voxel-level localization and expert-annotated CoT rationales, intended to imbue VLMs with robust grounded reasoning abilities for this anatomical region.

\paragraph{General Vision Language Models with Chain-of-Thought Reasoning.}
The integration of CoT reasoning into vision-language models has emerged as a critical advancement for complex visual understanding tasks. LLaVA-CoT~\cite{xu2024llava} pioneered a framework for teaching VLMs explicit multi-step reasoning, demonstrating significant performance improvements on multimodal reasoning benchmarks.
GCoT~\cite{wu2025grounded} introduces grounded chain-of-thought reasoning, explicitly connecting visual regions to reasoning steps, enhancing both accuracy and interpretability in complex visual tasks.
Other notable advances include Visual CoT~\cite{shao2024visual}, which proposes a multi-turn processing pipeline
that dynamically focuses on visual inputs and provides interpretable thought, and MM-CoT~\cite{zhang2023multimodal}, which incorporates language and vision modalities into a two-stage framework that separates rationale generation and
answer inference.
SpatialRGPT~\cite{cheng2024spatialrgpt} and RegionGPT~\cite{guo2024regiongpt} further enhance reasoning capabilities by grounding language outputs to specific visual regions. These advancements collectively demonstrate that explicit reasoning pathways significantly improve VLMs' performance on tasks requiring fine-grained visual analysis and multistep logical deduction.  OpenAI's o3 and o4-mini models~\cite{openai2025o3o4} have showcased remarkable multimodal reasoning capabilities, integrating images directly into their chain of thought. 
However, vision-language reasoning in the medical domain remains to be explored.

\section{Dataset and Benchmark}

\subsection{Problem Formulation: 3D Grounded Clinical Reasoning}

\begin{figure}[!t]
    \centering
    \includegraphics[width=0.9\linewidth]{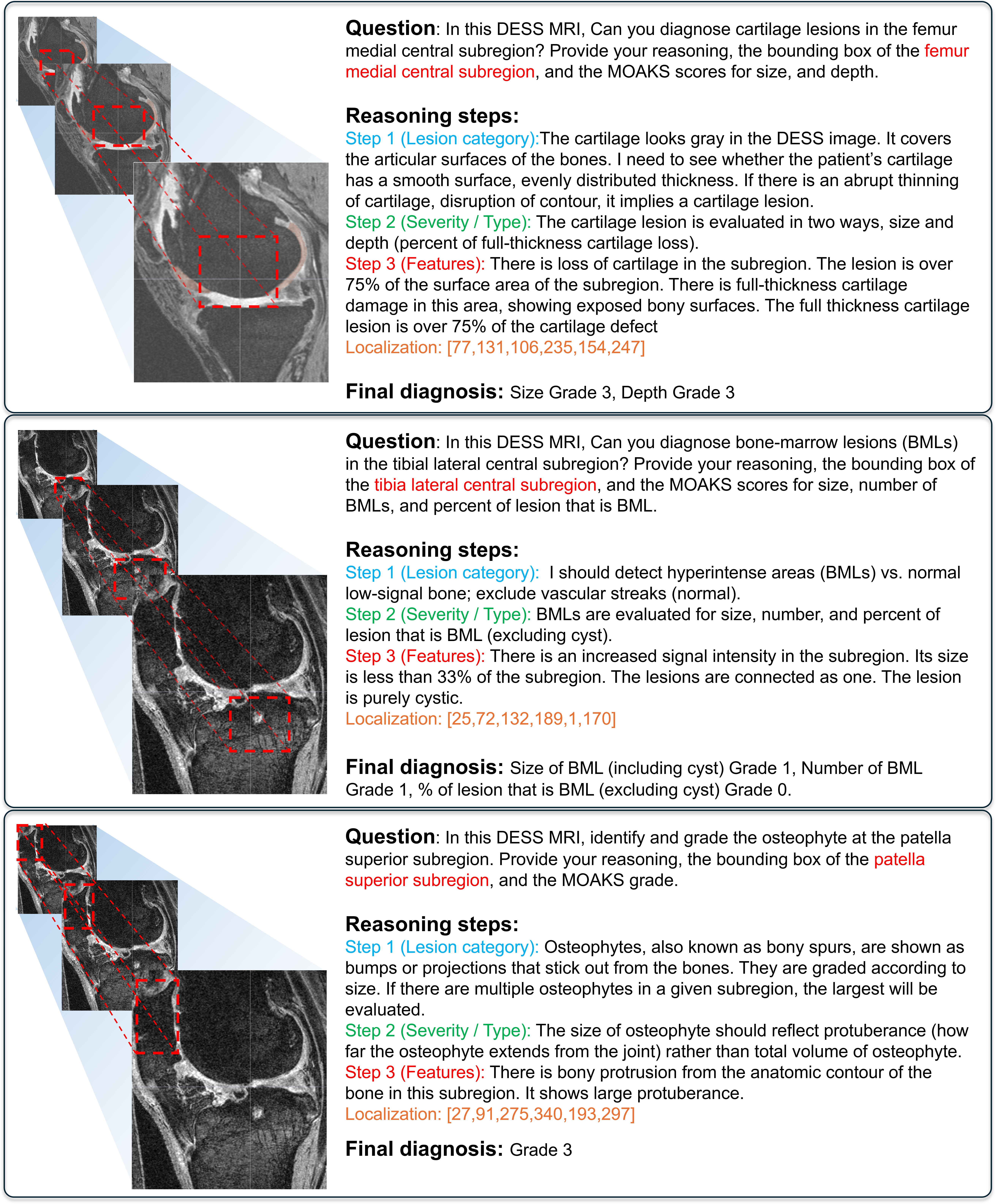}\vspace{-5pt}
    \caption{Example of 3DReasonKnee with Chain-of-Thought. }
    \label{fig:example}
\end{figure}

We formulate the problem as follows: Given a 3D knee MRI volume and a diagnostic question, the model must (1) localize the relevant anatomical region, (2) reason about abnormalities in that region, and (3) output a structured diagnosis that mirrors the MRI Osteoarthritis Knee Score (MOAKS) workflow. An example of quintuples can be found in Figure~\ref{fig:example}.



Each quintuple is represented as $\{I, Q, B, C, D\}$, where $I$ is the 3D MRI volume, $Q$ is the diagnostic question, $B$ is the 3D bounding box, $C$ is the CoT reasoning, and $D$ is the set of diagnostic severity grades.
This 3D knee MRI analysis framework processes the volume $I$ and diagnostic question $Q$ about a specific anatomical region and a single lesion category, then produces three outputs: a 3D bounding box $B$ that precisely localizes the relevant region, a concise Chain of Thought reasoning chain $C$ that transparently explains the analysis process, and a set of diagnostic severity grades $D$ that directly answers the original question for that lesion category. For a given anatomical region and a target lesion category $l$, this category is associated with a specific set of attributes: $\{a_1, \dots, a_m\}$. Each of these attributes has its own unique space of possible grades: $\{\mathcal{G}_{l}^{a_1}, \dots, \mathcal{G}_{l}^{a_m}\}$. The structured diagnosis $D$ is then represented as a set of severity grades, one for each attribute within that lesion category: $D = \{g_{l}^{a_1}, \dots, g_{l}^{a_m}\}$. Here, each grade $g_{l}^{a_j}$ is selected from the corresponding unique grade space $\mathcal{G}_{l}^{a_j}$ of the attribute $a_j$ for the lesion category $l$.

The model $\mathcal{M}$ predicts these three elements simultaneously, $(\hat B,\hat C,\hat D)=\mathcal{M}(I,Q)$,
providing a structured diagnosis that mirrors the MOAKS workflow by first identifying the region of interest, then reasoning about the specific lesion category within that region, and finally outputting the predicted severity grade for each attribute associated with that category, drawing from its unique grade space.

\subsection{\DATASET~ Construction Pipeline}
\begin{figure}[!th]
    \centering  
    \includegraphics[width=1\linewidth]{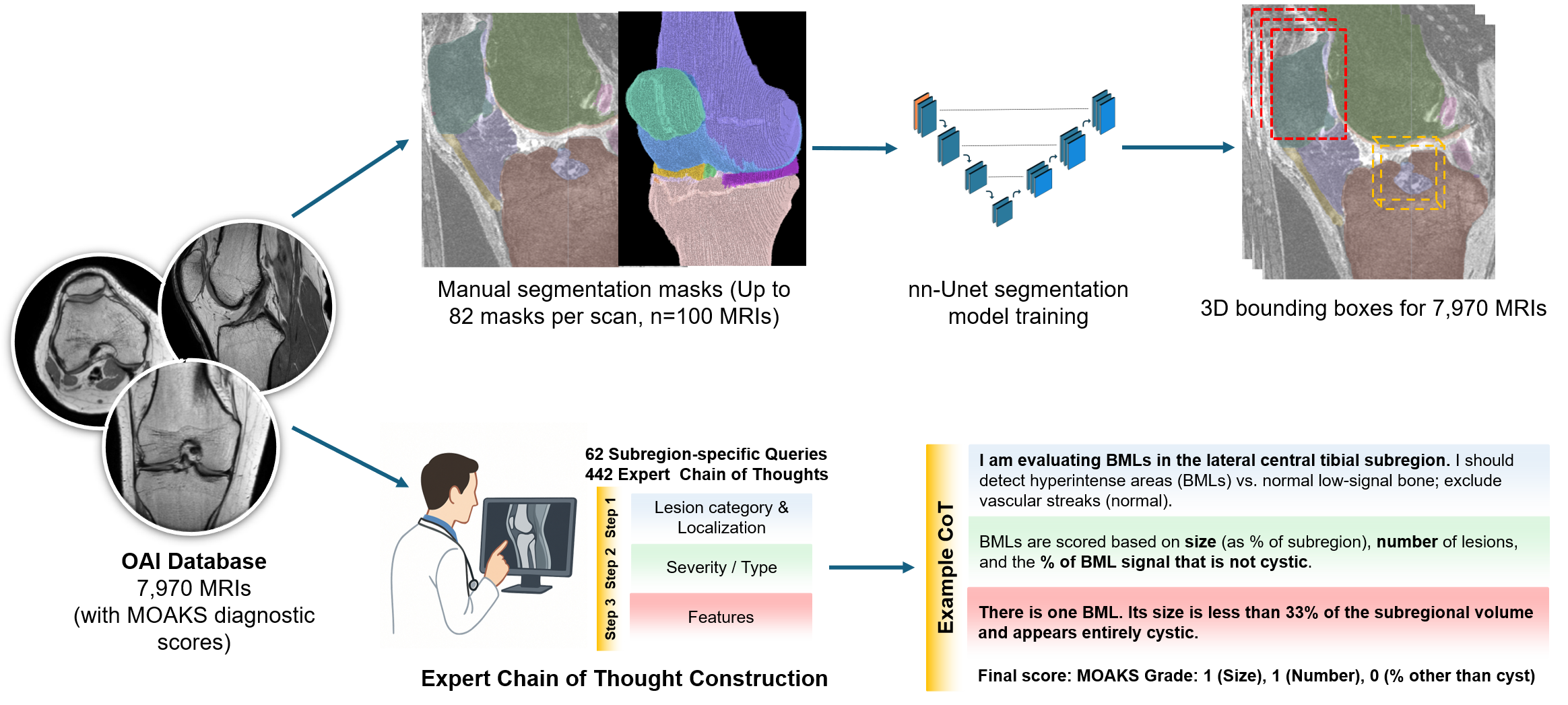}
    \caption{Overview of the 3DReasonKnee Dataset Construction Pipeline. MRIs (n=100) underwent manual segmentation to train an nn-Unet model \cite{isensee2021nnu}, enabling automatic generation of 3D bounding boxes for all scans (n=7,970). Clinicians developed subregion-specific queries and structured CoT based on the MOAKS system. }
    \label{fig:moaks}
\end{figure}

Our quintuplet dataset construction follows a systematic pipeline to create a comprehensive resource for 3D grounded clinical reasoning (see Figure~\ref{fig:moaks}).

\noindent \textbf{Data Source.} \DATASET~ builds upon the Osteoarthritis Initiative (OAI)~\cite{nevitt2006osteoarthritis}, a NIH-sponsored longitudinal study with 4,796 subjects (58\% female; age range: 45–79 years; mean age: 61.4 years) and over 45,000 MRI scans (https://nda.nih.gov/oai). We utilize 7,970 Double Echo Steady State (DESS) MRI sequences collected from baseline through 48-month follow-up, providing high-resolution volumetric data (160 slices per scan).  These scans include ground truth scores included in the OAI using the MOAKS framework~\cite{hunter2011evolution}, a comprehensive multi-label, multi-region scoring system for standardized knee osteoarthritis assessment.
MOAKS evaluates: bone marrow lesions (15 subregions), cartilage lesions (14 subregions), osteophytes (16 subregions), meniscal damage (6 subregions), ligaments/tendons (3 structures), hoffa-synovitis, effusion-synovitis, and periarticular features.

\noindent \textbf{3D Bounding Box Generation.} Two board-certified radiologists and one orthopedic surgeon performed manual annotation of anatomical subregions using a specialized 3D annotation tool. Each scan required approximately \textbf{4.5 expert hours} for comprehensive annotation, resulting in a substantial time investment of over \textbf{450 hours} for the complete high-fidelity subset of 100 scans.  We split the manual annotation data and used the standard nnU-Net pipeline~\cite{isensee2021nnu}  to build a segmentation model for all anatomical structures, subregions, and lesions. We then applied the trained nnU-Net to generate pseudo labels for all MRI scans with expert-adjudicated MOAKS assessments. As most of the current vision-language models primarily support bounding boxes rather than segmentation masks~\cite{bai2025qwen2,bannur2024maira}, we extracted the largest connected component for each subregion from the model's output and generated axis-aligned 3D bounding boxes. This process created a comprehensive dataset where each scan is linked with its corresponding MOAKS and relevant anatomical subregions.

\noindent \textbf{Expert Question Development \& Chain-of-Thought Generation.} A board-certified orthopedic surgeon with over seven years of experience in knee surgery developed a comprehensive set of 62 subregion-specific diagnostic queries, each aligned with distinct MOAKS components. For each targeted lesion and anatomical subregion combination, the clinician carefully formulated structured questions designed to prompt precise diagnostic evaluations. Subsequently, detailed step-by-step CoT were crafted to emulate the clinical reasoning pathway by clinicians. Specifically, the clinician first outlined the key pathological characteristics of each lesion type as observed on DESS MRI, emphasizing distinctive imaging features and points for distinguishing pathology from normal anatomy. In alignment with MOAKS criteria, the clinician then systematically translated these into reasoning steps, clearly articulating the conditions and rationale behind each  grade.


\noindent \textbf{Dataset Integration.} We linked the 3D bounding box annotations with corresponding expert queries, reasoning chains, and MOAKS ground truth labels to create a comprehensive multimodal dataset enabling both localization and reasoning tasks.

\subsection{\BENCHMARK{}: Evaluating 3D Multimodal Grounded Clinical Reasoning}

Our benchmark, \BENCHMARK{}, comprehensively evaluates a vision-language model's ability to perform 3D multimodal grounded clinical reasoning for knee osteoarthritis MRI assessment, assessing both diagnostic accuracy and anatomical localization as detailed in the following:

\begin{enumerate}
    \item \textbf{Final Diagnostic Accuracy:} We assess the accuracy of predicted MOAKS severity grades, grouped into 7 major categories: Bone marrow lesions (BML), Cartilages, Osteophytes, Menisci, Ligaments \& Tendons, Synovitis \& Effusion, and Periarticular Features. We report classification accuracy for each.
    \item \textbf{Anatomical Subregion Localization:} We evaluate the accuracy of predicted 3D bounding boxes using the 3D Intersection over Union (IoU):
       $$IoU_{3D} = \frac{Volume(B_p \cap B_{gt})}{Volume(B_p \cup B_{gt})},$$
       where $B_p$ is the predicted and $B_{gt}$ is the ground truth bounding box.
\end{enumerate}

By evaluating both localization and diagnostic accuracy, \BENCHMARK{} provides a holistic assessment of a VLM's grounded reasoning ability -- identifying the relevant 3D region and interpreting its visual information to reach a clinical diagnosis. The results in Table~\ref{tab:grading_results} and Table~\ref{tab:localization_results} offer an initial evaluation of existing models on this benchmark.

\begin{figure}[!th]
    \centering
    \includegraphics[width=\linewidth]{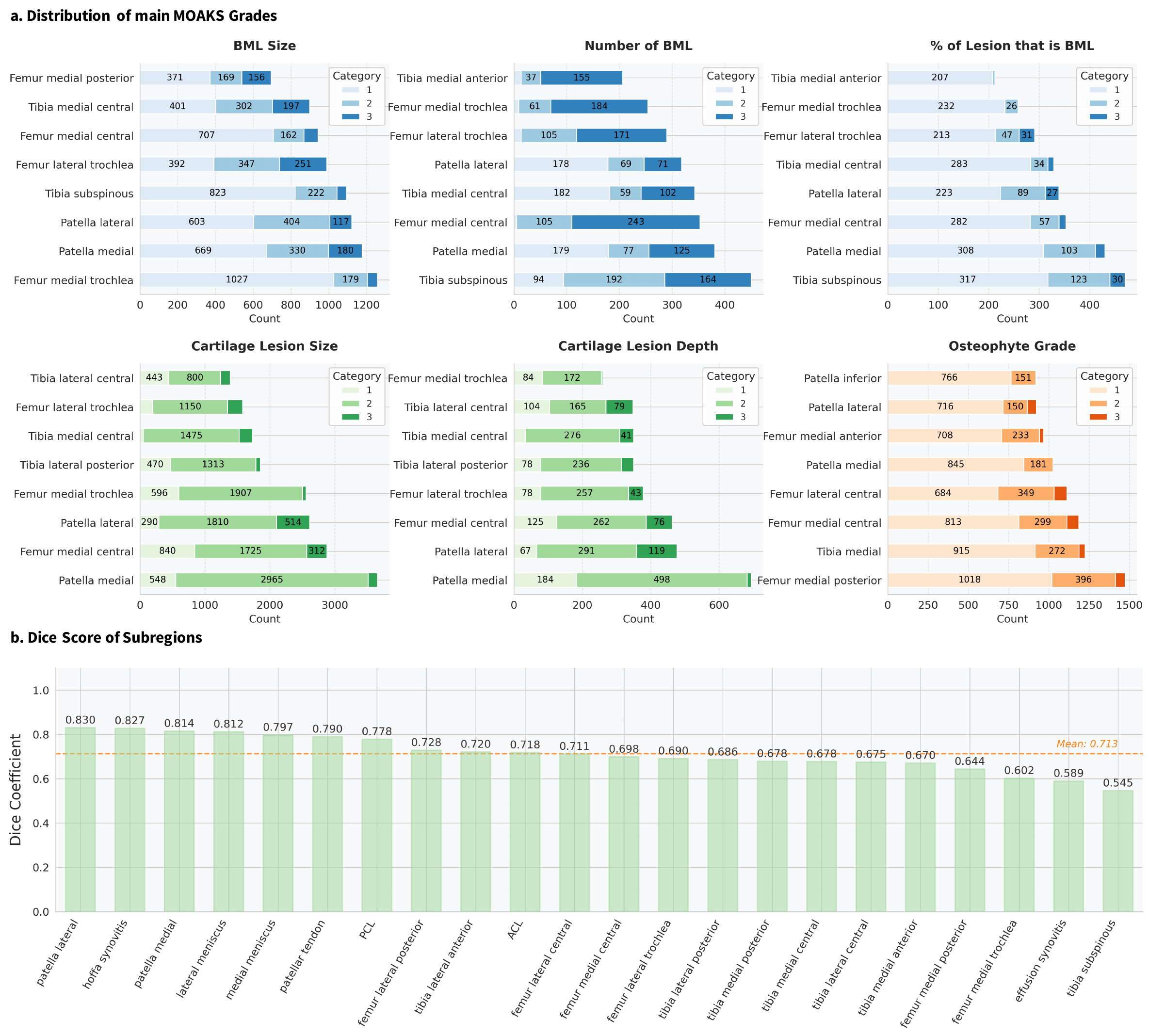} \vspace{-20pt}
    \caption{Dataset distribution statistics. (a) Distribution of main MOAKS grades across different anatomical regions. The stacked bars indicate the frequency of different severity grades (1-3) for each anatomical location, with the remaining proportion representing grade 0.
    (b) Dice score of the nnU-Net model of all subregions.
    }
    \label{fig:data_stat_1}
\end{figure}


\subsection{Dataset Analysis}

\noindent \textbf{Data Split.} Our data set comprises 7,970 high-resolution DESS MRI scans split into training (5,977 scans, 75\%), validation (797 scans, 10\%), and test sets (1,196 scans, 15\%) with no patient overlap. 

\noindent \textbf{Distribution of MOAKS Grades.} The dataset exhibits natural clinical variation in the severity of the pathology in different anatomical regions (Figure~\ref{fig:data_stat_1}a).
\begin{itemize}
    \item \textbf{BML Size, Number, \%}: The distribution indicates specific anatomical regions prone to developing BMLs, highlighting their relationship with OA progression. These regions include the medial aspect of the femur, the patella, and the area spanning from the tibia subspinous to the tibia medial central subregion.  
    \item \textbf{Cartilage Lesion Size and Depth}: Cartilage damage is most pronounced in the patella and the femur medial central subregion, reflecting anatomical regions most susceptible to OA-related cartilage loss, with disproportionate involvement of specific articular surfaces. 
    \item \textbf{Osteophyte Grade}: Osteophytes are more frequently observed in the medial compartment of the knee—specifically the femur medial, tibia medial, and patella medial subregions—reflecting the higher prevalence of medial knee OA in the population. 
\end{itemize}

\noindent \textbf{3D Bounding Box Annotations.} The 3D bounding box annotations provide precise spatial localization of relevant anatomical structures with comprehensive coverage across femoral, tibial, patellar, ligaments, and meniscal regions. 
The quality of these annotations is validated through Dice coefficient measurements against expert manual segmentations (Figure~\ref{fig:data_stat_1}b). The mean Dice score of 0.717 across all regions indicates high precision in localization. The bounding boxes for subregions showed particularly strong performance in the lateral patella (0.830), the hoffa synovitis (0.827), and the medial patella (0.814) subregions. Even the worst performing subregion, such as the tibial subspinous (0.545) exceed acceptable thresholds for clinical relevance. Details can be found here \href{https://huggingface.co/datasets/rajpurkarlab/3DReasonKnee}{\textcolor{blue}{(rajpurkarlab/3DReasonKnee)}}.

\section{Experiments and Baselines}

\noindent \textbf{Experimental Setup.} We benchmarked various baseline models using \DATASET. 
Our selection spans specialized 3D Medical VLMs (Med3DVLM~\cite{xin2025med3dvlm}), general-purpose VLMs with video input capabilities (Qwen2.5-VL-3B-Instruct, Qwen2.5-VL-7B-Instruct~\cite{bai2025qwen2}), and state-of-the-art API-based models (o1, using the ``2024-12-01-preview'' version via Azure). Implementation details are available at \href{https://huggingface.co/datasets/rajpurkarlab/3DReasonKnee}{\textcolor{blue}{rajpurkarlab/3DReasonKnee}}.

\noindent \textbf{Experimental Settings.} We evaluated the models under three different settings: zero-shot, zero-shot with instruction schema, and supervised fine-tuning, which allows us to quantify the inherent difficulty of the proposed tasks for them.
\begin{itemize}
    \item \textbf{Zero-shot Inference.} For our zero-shot evaluation, we provide each model with the 3D MRI scan and a question prompt without any examples of the expected reasoning chain. 
    For example, \textit{``In this DESS MRI, identify and grade the osteophyte at the patella lateral subregion. Provide your reasoning, and the MOAKS grade.''}
    The models rely solely on their pre-trained knowledge to analyze the 3D medical images, identify relevant anatomical structures, and generate appropriate clinical reasoning and MOAKS scores.
    \item \textbf{Zero-shot Inference with Instruction Schema.} In this enhanced zero-shot setting, we augment the base prompt with a structured instruction schema that outlines the expected reasoning process. 
    For example, \textit{``In this DESS MRI, identify and grade the osteophyte at the patella lateral subregion. Osteophytes are bony outgrowths at joint margins. Osteophytes are scored using MOAKS grading: Grade 0: None, Grade 1: Small, Grade 2: Medium, Grade 3: Large. If there are multiple osteophytes in a given subregion, the largest will be evaluated. The size of osteophyte should reflect protuberance (how far the osteophyte extends from the joint) rather than total volume of osteophyte. Provide your reasoning and the MOAKS grade.''}. This instruction provides models with a systematic framework for interpreting 3D MRI data without prescribing specific findings, allowing us to evaluate how effectively models can apply structured reasoning processes to clinical assessment tasks.
    \item \textbf{Zero-shot Inference with Instruction Schema and Ground Truth Region.} In this setting, first, we augmented the zero-shot prompt with the structured instruction schema used in the previous baseline, explicitly outlining the expected reasoning process. Second, we cropped the input MRI volume to include only the ground-truth region of interest. By eliminating the need for the model to localize the relevant anatomical structure, this setting isolates its reasoning and grading capabilities.
    \item \textbf{Supervised Fine-tuning.} For the supervised approach, we fine-tune each baseline model on our training set. Models with the best performance on our validation set are used for evaluation. 
\end{itemize}

\begin{table}[ht]
\tbl{\small Results Summary of MOAKS Grading Performance. We report average accuracy on attributes for each lesion category (BML, Cartilage, Osteophyte, Meniscus, and Others) as well as the Overall average across all categories.}{\vspace{1pt}
\setlength{\tabcolsep}{3pt}     
\resizebox{\textwidth}{!}{
\begin{tabular}{llcccccc}
\toprule
\textbf{Model} & \textbf{Method} & \textbf{BML} & \textbf{Cartilage} & \textbf{Osteophyte} & \textbf{Meniscus} & \textbf{Others} & \textbf{Overall} \\
\midrule
Qwen2.5VL-3B & Zero-shot  & 0.081 & 0.253 & 0.251 & 0.093 & 0.159 & 0.158 \\
Qwen2.5VL-7B & Zero-shot  & 0.642 & 0.309 & 0.348 & 0.347 & 0.469 & 0.470 \\
o1 & Zero-shot  & 0.479 & 0.553 & 0.353 & 0.332 & 0.476 & 0.477 \\
\midrule
Qwen2.5VL-3B & Instruction Zero-shot & 0.092 & 0.149 & 0.253 & 0.205 & 0.146 & 0.146 \\
Qwen2.5VL-7B & Instruction Zero-shot & 0.629 & 0.466 & 0.342 & 0.297 & 0.503 & 0.504 \\
o1 & Instruction Zero-shot & 0.635 & 0.579 & 0.345 & 0.391 & 0.567 & 0.568 \\
\midrule
Qwen2.5VL-7B & \shortstack{Instruction Zero-shot w/ GT region} & 0.697 & 0.549 & 0.338 & 0.166 & 0.563 & 0.556 \\
\midrule
Qwen2.5VL-3B & SFT w/o CoT  & 0.706 & 0.602 & 0.345 & 0.387 & 0.616 & 0.613  \\
Med3DVLM & SFT w/o CoT  &  0.670 &  0.578 &  0.330 & 0.346 &  0.600  & 0.596\\
\midrule
Qwen2.5VL-3B & SFT w/ CoT  & 0.667 & 0.600 & 0.342 & 0.340 & 0.590 & 0.588 \\
Med3DVLM & SFT w/ CoT  & 0.671 & 0.579 & 0.333 & 0.343 & 0.605 & 0.581 \\
\bottomrule
\end{tabular}
}
}
\label{tab:grading_results}
\end{table}

\begin{table}[ht]
\centering
\tbl{Results Summary of Anatomical Subregion Localization (IoU). Values represent average IoU scores for each subregion of each anatomical structure across different models.}{\vspace{1pt}
\begin{tabular}{lccccccc}
\toprule
\textbf{Model} & \textbf{Patella} & \textbf{Femur} & \textbf{Tibia} & \textbf{Meniscus} &\textbf{ACL} & \textbf{PCL} & \textbf{Synovitis} \\
\midrule
Qwen2.5VL-3B  &  0.293 & 0.275 & 0.322 & 0.174 & 0.299 & 0.305 & 0.519 \\
Med3DVLM & 0.298 & 0.314 & 0.359 & 0.197 & 0.319 & 0.384 & 0.470 \\

\bottomrule
\end{tabular}
}
\label{tab:localization_results}
\end{table}

\paragraph{Baselines.} Table~\ref{tab:grading_results} summarizes the performance of different approaches on MOAKS component grading diagnostic accuracy, while Table~\ref{tab:localization_results} presents the localization performance across anatomical subregions.
In zero-shot settings, even state-of-the-art models struggle with the complex task of MOAKS grading, with Qwen2.5VL-3B achieving only 0.158 overall accuracy. The larger Qwen2.5VL-7B model shows some improvement (0.470), comparable to the o1 model (0.477), but both fall short of clinically acceptable performance. 
Note that medical VLMs like Med3DVLM fail to follow the instructions and give valid outputs. 
For models that successfully follow instructions, prompting with structured instruction schemas significantly improves performance across models. When prompted with the instruction schema and the ground truth region, we observed a 0.05 increase in overall accuracy (0.556 vs. 0.504) compared to the Instruction Zero-shot baseline. This indicates that when the localization step is solved, the model’s reasoning and final diagnostic grading improve.
The o1 model showed the most substantial gains from instruction following, with overall accuracy increasing from 0.477 to 0.568 (+19.1\%). This improvement was particularly pronounced for BML assessment, where accuracy increased from 0.479 to 0.635 (+15.6\%). Supervised fine-tuning (SFT) yielded substantial performance improvements, with Qwen2.5VL-3B achieving 0.613 overall accuracy. While incorporating CoT reasoning during this SFT phase did not translate to accuracy gains in our initial experiments, the supervised models still significantly outperformed o1. We believe that the rich, expert-generated CoT annotations within our dataset hold considerable potential that may be more effectively unlocked with alternative training paradigms beyond standard supervised fine-tuning. Additionally, we observed a clear positive correlation between localization accuracy (IoU) and diagnostic accuracy for our supervised models, with the Med3DVLM model trained with CoT exhibiting the strongest positive correlation between these two metrics.

\paragraph{Failure Modes.} 
Taken together, these two experiments provide an initial failure analysis across all the models evaluated.
Grounding is a major failure point. Performance improves notably when the ground-truth region is provided, indicating that many errors arise from incorrect localization. Reasoning and grading remain challenging for certain categories. Even when localization is provided, categories such as Meniscus and Osteophyte still show low accuracy, suggesting that fine-grained reasoning about subtle lesion patterns is also a limitation.

\section{Discussion}
Our work lays the crucial groundwork for training grounded reasoning models in medical imaging. By demonstrating the current limitations of VLMs in this domain,  \DATASET~provides a vital resource to catalyze the development of models capable of reasoning through complex 3D medical data in a manner that mirrors the structured, region-first approach of clinicians. \DATASET~ provides a foundation for developing interpretable AI tools that integrate into diagnostic workflows and improve patient care. Future research directions stemming from this work are manifold. We believe this dataset shows promise for exploring reinforcement learning (RL) approaches, which can guide VLMs to diagnose by emulating the expert clinical process. \cite{liu2025x, lai2025med, schmied2025llms}. Notably, \DATASET~ features vision-language diagnostic problems paired with CoT rationales and a structured response format, a combination conducive to effective RL training. This dataset could serve as an SFT layer, establishing coherent thinking patterns and output structures, in conjunction with advanced RL techniques that strengthen both reasoning performance and generalization, as demonstrated in successful approaches like OpenAI o3. Another promising training avenue involves exploring novel training processes that directly embed both the full image and the localized subregion/lesion information, a strategy that has been eplored in 2D but is currently limited by computational constraints for 3D images \cite{shao2024visual}.  Furthermore, the structured nature of \DATASET~opens doors for investigating longitudinal reasoning, analyzing disease progression at different time points, and ultimately contributing to prognosis and treatment planning. Extending this framework to other medical imaging modalities beyond knee MRI and other joints such as hand and hip OA represents another significant future direction, paving the way for more broadly applicable and clinically impactful grounded reasoning models. \cite{Haugen2015HOAMRIS, Roemer2011HOAMS}

\section{Acknowledgments}
This Research was supported by a grant of the Boston-Korea Innovative Research Project through the Korea Health Industry Development Institute(KHIDI), funded by the Ministry of Health \& Welfare, Republic of Korea(grant no. : RS-2024-00403047). The OAI is a public-private partnership comprised of five contracts (N01- AR-2-2258; N01-AR-2-2259; N01-AR-2- 2260; N01-AR-2-2261; N01-AR-2-2262) funded by the National Institutes of Health, a branch of the Department of Health and Human Services, and conducted by the OAI Study Investigators. Private funding partners include Merck Research Laboratories; Novartis Pharmaceuticals Corporation, GlaxoSmithKline; and Pfizer, Inc. Private sector funding for the OAI is managed by the Foundation for the National Institutes of Health.

\bibliographystyle{plain}
\bibliography{ws-pro-sample}

\clearpage

\end{document}